\documentclass{article}

\usepackage[utf8]{inputenc}
\usepackage[T1]{fontenc}
\usepackage{graphicx}
\usepackage{authblk}
\usepackage[breaklinks=true]{hyperref}
\usepackage{breakcites}
\usepackage{url}
\usepackage{xspace}

\usepackage{amsmath}
\usepackage{amssymb}

\usepackage[ucmark=true,hyperfirst=false]{glossaries}
\glsdisablehyper
\newacronym{1-D}{1-D}{1-Dimensional}
\newacronym{2-D}{2-D}{2-Dimensional}
\newacronym{3-D}{3-D}{3-Dimensional}
\newacronym{ADC}{ADC}{Analog to Digital Converter}
\newacronym{AGC}{AGC}{Automatic Gain Control}
\newacronym{AI}{AI}{Artificial Intelligence}
\newacronym{AMHA}{AMHA}{\`A Mon Humble Avis}
\newacronym{ARAIM}{ARAIM}{Advanced Receiver Autonomous Integrity Monitoring System}
\newacronym{ASIC}{ASIC}{Application-Specific Integrated Circuit}
\newacronym{AWGN}{AWGN}{Additive White Gaussian Noise}
\newacronym{BAW}{BAW}{Bulk Acoustic Wave}
\newacronym{BCE}{BCE}{Binary Cross Entropy}
\newacronym{BOC}{BOC}{Binary Offset Carrier}
\newacronym{BPSK}{BPSK}{Binary Phase Shift Keying}
\newacronym{CNN}{CNN}{Convolutional Neural Network}
\newacronym{CNN-HL}{CNN-HL}{CNN-Histogram Loss}
\newacronym{CNN-HL TG}{CNN-HL TG}{CNN-Histogram Loss Truncated Gaussian}
\newacronym{CNN-HL VM}{CNN-HL VM}{CNN-Histogram Loss Von Mises}
\newacronym{CNN-Reg}{CNN-Reg}{CNN-Regression}
\newacronym{CPW}{CPW}{Coplanar Wave Line}
\newacronym{CW}{CW}{Carrier Wave}
\newacronym{DME}{DME}{Distance Measuring Equipment}
\newacronym{DC}{DC}{Direct frequency Conversion}
\newacronym{DC2}{DC}{Direct Current}
\newacronym{DF}{DF}{Dual-Frequency}
\newacronym{DL}{DL}{Deep Learning} 
\newacronym{DLL}{DLL}{Delay-Locked Loop}
\newacronym{DNN}{DNN}{Deep Neural Network}
\newacronym{DS}{DS}{Direct Sampling}
\newacronym{DSP}{DSP}{Digital Signal Processor}
\newacronym{ENAC}{ENAC}{\'Ecole Nationale de l'Aviation Civile}
\newacronym{EMD}{EMD}{Earth Mover's Distance}
\newacronym{ENC}{ENC}{European Navigation Conference}
\newacronym{FLL}{FLL}{Frequency Locked Loop}
\newacronym{FIR}{FIR}{Finite Impulse Response}
\newacronym{FPGA}{FPGA}{Field-Programmable Gate Array}
\newacronym{GBAS}{GBAS}{Ground-Based Augmentation System}
\newacronym{GLONASS}{GLONASS}{Global Navigation Satellite System}
\newacronym{GNSS}{GNSS}{Global Navigation Satellite System}
\newacronym{GPL}{GPL}{General Public License}
\newacronym{GPS}{GPS}{Global Positioning System}
\newacronym{GSL}{GSL}{GNU Scientific Library}
\newacronym{HL}{HL}{Histogram Loss}
\newacronym{ICD}{ICD}{Interface Control Document}
\newacronym{IF}{IF}{Intermediate Frequency}
\newacronym{iid}{i.i.d.}{Independent and Identically Distributed}
\newacronym{IIR}{IIR}{Infinite Impulse Response}
\newacronym{INS}{INS}{Inertial Navigation System}
\newacronym{ION}{ION}{Institute Of Navigation}
\newacronym{IRNSS}{IRNSS}{Indian Regional Navigational Satellite System}
\newacronym{KL}{KL}{Kullback-Leibler}
\newacronym{LEO}{LEO}{Low Earth Orbit}
\newacronym{LNA}{LNA}{Low-Noise Amplifier}
\newacronym{LOS}{LOS}{Line-Of-Sight}
\newacronym{LSTM}{LSTM}{Long Short-Term Memory}
\newacronym{MAE}{MAE}{Mean Absolute Error}
\newacronym{MEDLL}{MEDLL}{Multipath Estimating Delay Lock Loop}
\newacronym{ML}{ML}{Machine Learning}
\newacronym{MLP}{MLP}{Multi-Layer Perceptron}                                              
\newacronym{MOPS}{MOPS}{Minimum Operational Performance Specification}
\newacronym{MSB}{MSB}{Most Significant Bit}
\newacronym{NBI}{NBI}{Narrow Band Interference}
\newacronym{NCO}{NCO}{Numerically Controlled Oscillator}
\newacronym{NB}{NB}{Narrow Band}
\newacronym{NLOS}{NLOS}{Non Line-Of-Sight}
\newacronym{NN}{NN}{Neural Networks}
\newacronym{OCXO}{OCXO}{Oven Controlled Crystal Oscillator}
\newacronym{oktalse}{Oktal-SE}{Oktal-Synthetic Environment}
\newacronym{OS}{OS}{Open Service}
\newacronym{PCB}{PCB}{Printed Circuit Board}
\newacronym{PDF}{PDF}{Probability Density Function}
\newacronym{PhD}{PhD}{Philosophi\ae Doctor}
\newacronym{PLL}{PLL}{Phase-Locked Loop}
\newacronym{PPS}{PPS}{Pulse Per Second}
\newacronym{PRN}{PRN}{Pseudo-Random Noise}
\newacronym{PSD}{PSD}{Power Spectral Density}
\newacronym{PVT}{PVT}{Position Time Velocity}
\newacronym{POT}{POT}{Python Optimal Transport}
\newacronym{QPSK}{QPSK}{Quadrature Phase Shift Keying}
\newacronym{QZSS}{QZSS}{Quasi-Zenith Satellite System}
\newacronym{RELU}{RELU}{Rectified Linear Unit}
\newacronym{RF}{RF}{Radio Frequency}
\newacronym{RGB}{RGB}{Red Green Blue}
\newacronym{RHCP}{RHCP}{Right Hand Circular Polarization}
\newacronym{RMS}{RMS}{Root Mean Square}
\newacronym{RNN}{RNN}{Recurrent Neural Network}
\newacronym{RNSS}{RNSS}{Radio Navigation Satellite Service}
\newacronym{SAW}{SAW}{Surface Acoustic Wave}
\newacronym{SBAS}{SBAS}{Satellite-Based Augmentation System}
\newacronym{SDR}{SDR}{Software-Defined Radio}
\newacronym{SIS}{SIS}{Signal In Space}
\newacronym{SMA}{SMA}{SubMiniature version A}
\newacronym{SSL}{SSL}{Semi-Supervised Learning}
\newacronym{SV}{SV}{Space Vehicle}
\newacronym{SVM}{SVM}{Support Vector Machines}
\newacronym{TCXO}{TCXO}{Temperature Controlled Crystal Oscillator}
\newacronym{THD}{THD}{Total Harmonic Distortion}
\newacronym{TIE}{TIE}{Time Interval Error}
\newacronym{VGA}{VGA}{Variable Gain Amplifier}
\newacronym{VGG}{VGG}{Visual Geometry Group}
\newacronym{VSWR}{VSWR}{Voltage Standing Wave Ratio}
\usepackage[thinqspace, squaren, Gray, cdot]{SIunits}

\usepackage{calc}

\usepackage{textcomp}

\usepackage{subcaption}

\DeclareGraphicsExtensions{.eps,.pdf,.png}

\newcommand{\dBHz}{\ensuremath{\deci\bel\hertz}\xspace}

\newcommand{\RR}{\mathcal{R}}

\newcommand{\CNZERO}{\ensuremath{\mathit{C\!/\!N0}}\xspace}
\newcommand{\dsp}{\displaystyle}

\newcommand{\ignore}[1]{}

\providecommand{\keywords}[1]
{
  \small	
  \textbf{\textit{Keywords---}} #1
}

\title{Wasserstein distance based semi-supervised manifold learning and application to GNSS multi-path detection}
\author[1]{Antoine Blais \thanks{Corresponding author: \texttt{antoine.blais@recherche.enac.fr}}}
\author[1,2]{Nicolas Couëllan \thanks{\texttt{nicolas.couellan@recherche.enac.fr}}}
\affil[1]{ENAC, Université de Toulouse, 7 Avenue Édouard Belin, \protect \\BP 54005, 31055 Toulouse Cedex 4, France}
\affil[2]{Institut de Mathématiques de Toulouse, Université de Toulouse, UPS IMT, F-31062 Toulouse Cedex 9, France}

\begin{document}

\maketitle

\abstract{The main objective of this study is to propose an optimal transport based semi-supervised approach to learn from scarce labelled image data using deep convolutional networks. The principle lies in implicit graph-based transductive semi-supervised learning where the similarity metric between image samples is the Wasserstein distance. This metric is used in the label propagation mechanism during learning. We apply and demonstrate the effectiveness of the method on a \gls{GNSS} real life application. More specifically, we address the problem of multi-path interference detection. Experiments are conducted under various signal conditions. The results show that for specific choices of hyperparameters controlling the amount of semi-supervision and the level of sensitivity to the metric, the classification accuracy can be significantly improved over the fully supervised training method.}


\section{Introduction}
Many engineering application contexts could take advantage of recent and successful developments of artificial intelligence. However, it is often difficult to gather a sufficiently large set of labelled samples to train models such as deep learning architectures. This is particularly true for \gls{CNN} based algorithms characterised by a large amount of convolutional blocks containing numerous filters. Indeed, the more complex the architecture, the more parameters there are to be trained, requiring therefore more labelled data. 

The principle of \gls{SSL} has been introduced to handle situations where large amount of data is available but only few samples are labelled \cite{chapelle2006, vanEngelen2019ASO, 10.1109/TPAMI.2022.3201576}. There are two main categories of \gls{SSL} techniques. First, inductive methods such as wrappers method \cite{zhu2008} or co-training methods \cite{blum1998, xu2013} have been proposed. They construct pseudo-labels using a pretrained supervised learning procedure and require usually a second training phase with the pseudo-labels. When a similarity measure between samples is available, the second class of techniques known as transductive methods can also be employed \cite{Weston2012}. The graph-based methods are popular approaches along this line. They rely on the reasonable assumption that in practice data lie on a low dimensional smooth manifold \cite{chapelle2006}. In other words, it assumes that when data are similar their labels should be similar. The similarity metric between data allows the construction of an explicit or implicit graph between the samples. In graph-based transductive \gls{SSL} methods, the aim is to train a model on both labelled and unlabelled data through a label inheritance mechanism between similar samples, also known as label propagation. 

When dealing with image input samples, it has been shown that the \gls{EMD} and more generally the Wasserstein distance \cite{Friesecke2025} is a relevant similarity measure that matches perceptual similarity better than other distances commonly used in image processing \cite{rubner2000}. Therefore, the availability of an efficient similarity measure for image samples advocates for the use of a graph-based SSL approach to train deep learning models when labelled images are scarce. \gls{SSL} has been proposed in the past in the context of neural networks~\cite{Weston2012} and the Wasserstein distance has been utilized in other deep learning contexts such as Wasserstein GAN \cite{pmlr-v70-arjovsky17a} or Wasserstein distance based deep adversarial transfer learning \cite{CHENG202035}. However, to the best of our knowledge, it has not been proposed for a real life application involving semi-supervised \gls{CNN} in the literature. The aim of this study is to show that using a similarity measure based on optimal transport, image classification with few labelled samples can be significantly improved using \gls{SSL}. The real life application we consider is the detection of interference in \gls{GNSS} signals. The signals are shaped as multi-channel images where the similarity measure between signals is defined as the Wasserstein distance metric. The detection of multi-path interference is of great importance for \gls{GNSS} as their presence can greatly alter the estimated position of the receiver. When receivers are embedded in systems such as aircraft, providing efficient and robust \gls{GNSS} is critical.

The main contribution of this study is the original combination of \gls{SSL} and \gls{CNN} using the Wasserstein distance metric. More specifically, to the best of our knowledge this is one of the first study demonstrating a successful implementation of such ideas to improve predicting performance of a \gls{CNN} through \gls{SSL}. Additionally, our implementation preserves the decomposition structure of the \gls{SSL} regularised training loss in order to process large unlabelled data sets by batches of data and can then benefit from stochastic gradient based techniques during training. Furthermore, the use of optimal transport to evaluate the similarity between \gls{GNSS} signals has never been proposed before. This work also shows that in \gls{GNSS} applications where limited labelled data are available, one can use the regularity assumption to train \gls{ML} models. Least but not last, this is a step towards data frugality in deep learning.

The rest of this article is structured as follows. First, section~\ref{section_ssl} clarifies the principle of \gls{SSL} and the underlying concept of smoothness enforcing regularization. Then, the \gls{GNSS} application is presented in section~\ref{section_gnss}. At the same time, the distance metric chosen for semi-supervised training is exposed as it is data dependant. Section~\ref{section_experiments} details afterwards the test bench and the experiments which have been set up to assess the performance of \gls{SSL} in the context of \gls{GNSS} multi-path detection. Next, results are presented and thoroughly discussed in section~\ref{section_results}. Finally, section \ref{section_conclusions} concludes the article.

\section{Semi-supervised learning}
\label{section_ssl}

In this section, we introduce the general model that makes use of general data hypothesis in order to learn from unlabelled data. Next, we specify the choices we propose to construct the numerical model and explain the computation of distances between data samples.

\subsection{General semi-supervised framework}
When learning from data, several underlying assumptions are commonly admitted \cite{chapelle2006}. The \textit{manifold hypothesis} is a basic assumption in most machine learning problems. It states that the high dimensional data must lie in low dimensional manifold thus avoiding the curse of dimensionality when designing learning algorithms. The \textit{smoothness assumption} is also commonly admitted and is of great importance in the model we consider. It assumes that the label function of the data varies smoothly with the distance between samples and even more so when in high density regions of the data space.\\
Consider a neural network $f(x,\theta)$ where $x\in\mathcal{D}:=\mathcal{D}_S\cup\mathcal{D}_U$ where $\mathcal{D}_S$ is the set of all labelled samples (later $\mathcal{Y}$ will denote the corresponding set of labels), $\mathcal{D}_U$ is the set of unlabelled samples,  
and $\theta$ are the neural network parameters (weights and biases). As in \cite{Weston2012}, the above smoothness assumption may be used as a regularizer in the empirical risk minimization training problem in order to train the neural network. More specifically, the following composite loss function may be constructed:
\begin{equation}
\dsp \min_\theta \sum_{(x,y)\in\mathcal{D}_S\times\mathcal{Y}} \mathcal{L}_S(y,f(x,\theta)) + \lambda \sum_{(x,x')\in \mathcal{D}\times\mathcal{D}} \mathcal{L}_U(f(x,\theta),f(x',\theta))   
\end{equation}
where $\mathcal{L}_S$ and $\mathcal{L}_U$ are respectively loss and regularization functions for supervised and unsupervised learning respectively, $y\in\mathcal{Y}$ are the labels, and $\lambda$ is a hyperparameter that controls the trade-off between supervision $\mathcal{L}_S$ and the smoothness enforcing regularization $\mathcal{L}_U$. The loss  $\mathcal{L}_S$ may be any suitable error function for the learning problem at hand that calculates discrepancies between label and model prediction (ex: Mean squared error, Mean absolute error, binary cross-entropy,...). To choose $\mathcal{L}_U$, several strategies have been proposed \cite{Weston2012}. Our choice will be explained next.

\subsection{Smoothness enforcing regularization}
To enforce smoothness of the model so that unlabelled data can inherit information from labelled neighbours, we propose to carry out label propagation through the model \cite{chapelle2006}. The idea is to calculate model predictions for pairs of data samples and weight the corresponding squared difference between the pair of predictions by their corresponding relative distance in the input space. The choice of distance metric in the input space to decide whether input samples are neighbours or not is specific to the nature of data and will be addressed later in Section~\ref{section_Wasserstein}. More specifically, our choice for $\mathcal{L}_U$ is defined as follows:
consider that $\{1,\ldots,L\}$ contains the indices of labelled input samples and $\{L+1,\ldots,L + U\}$ contains the indices of unlabelled input samples (with usually $U \gg L$), we define the total smoothing regularization as
\begin{equation}
\dsp \sum_{\substack{{i, j=1} \\ (i \le L) \oplus (j \le L)}}^{L+U}
\mathcal{L}_U(f(x_i,\theta),f(x_j;\theta))\label{LU}
\end{equation}
with
\begin{equation}
\dsp \mathcal{L}_U(f(x_i,\theta),f(x_j,\theta))=W_{ij}\|f(x_i,\theta)-f(x_j,\theta)\|^2
\end{equation}
and where $W_{ij}\in[0,1]$ is a weight parameter that is close to $1$ if the distance between $x_i$ and $x_j$ is small and near $0$ otherwise. In other words, $W_{ij}$ controls the similarity between $x_i$ and $x_j$ and is computed through the Gaussian kernel similarity measure as follows~\cite{Belkin2003}:
\begin{equation}
W_{ij}=\exp{\left\{-\frac{d(x_i,x_j)^2}{\sigma}\right\}}
\end{equation}
where $d(x_i,x_j)$ is the distance metric between the samples $x_i$ and $x_j$ (see Section~\ref{section_Wasserstein}) and $\sigma$ is a bandwidth parameter to control the sensitivity of $W_{ij}$ to the similarity within the sample pair.
Observe that in (\ref{LU}), the sum over $i$ and $j$ is taken over all pairs of samples once and that the operator $\oplus$ indicates that a pair where both samples have labels is excluded from the sum. This specific handling of pairs of samples ensures that we are not computing any smoothing term when both samples have labels as this might be a conflicting objective for two samples that have distinct labels . Indeed, this mechanism gives more importance to ground truth through the $\mathcal{L}_S$ error term rather than relying on the smoothness assumption.

When training the neural network with a stochastic optimization algorithm such as stochastic gradient descent (\texttt{SGD})  or its variants (ex: \texttt{ADAM}, \texttt{ADAGRAD} and other) \cite{Wright_Recht_2022,kingma2017}, it is important that the loss function remains separable with respect to samples. Note that the way we define $\mathcal{L}_U$ preserves this separability and it is possible to construct mini-batches where only pairs of samples within the mini-batch are considered during the iterative process. Therefore the proposed procedure also scales up to large data sets.

\section{GNSS application}
\label{section_gnss}
In this section, we first describe the data used to assess the combination of \gls{CNN} and \gls{SSL}. The objective to detect samples contaminated by multi-path signals is also clarified. Then, the definition of the Wasserstein distance metric selected to effectively put into practice semi-supervised learning is given. The strength of this particular metric to perform image classification in our context is explained as well.

\subsection{GNSS Image data set}
The data of interest in this study are produced by the \gls{RF} front-end of a \gls{GNSS} receiver. More precisely, the signal available at the output of the \gls{RF} front-end comes in two orthogonal channels. It is then shaped into two \gls{2-D} matrices or equivalently a pair of images~\cite{Blais2022},~\cite{Gonzalez2024}. The components of the pair are noted $I$ and $Q$ by convention in the \gls{GNSS} community.  Figure~\ref{fig:I_Q_flat}(a) displays an example of a pair of images. Figure~\ref{fig:I_Q_3D}(a) in Appendix~\ref{appendix_I_Q_3D} offers an equivalent \gls{3-D} representation. Each pair of images is a snapshot of the received signal at a given time, a mix of the signal of interest, of the receiver noise and of any other disturbing interference captured by the antenna. A pair of images could be seen by analogy as the stereoscopic vision of a scene from a \gls{3-D} movie provided by polarized glasses.

This image representation has proven to be efficient when coupled to a \gls{CNN} to detect a multi-path, or echo, signal affecting the \gls{LOS} signal \cite{Blais2022}. It has also been demonstrated that the same association was effective to estimate the parameters of the multi-path signal~\cite{Gonzalez2024}. The \gls{CNN} implemented in the publications mentioned above need large amount of labelled data to be trained. However, real labelled data are scarce or expensive in the \gls{GNSS} field. Then, a synthetic data generator has been used to create training and testing data sets. In this study, the synthetic data generator~\cite{Blais2025a} is also used but the purpose is to show that it is possible to reduce the dependency to the generator by using semi-supervised learning. The objective is to analyze whether the performance in detecting multi-path contamination of the signal of interest is improved or not by introducing unlabelled data in the learning process. The source code of the synthetic data generator is freely available under the terms of the GNU \gls{GPL} version 3 license.

The original resolution of the images produced by the generator is 89x81 for physical compliance with the receiver hardware and signal processing requirements. However, previous research works have shown that a reduction in resolution to as low as 26x26 still gives high detection accuracy for a much lower training time and computing power. Hence, this resolution has been retained for this study.

Images are always composed of at least the \gls{LOS} signal and of the receiver noise. The \gls{LOS} signal is modelled as depicted in Figure~\ref{fig:I_Q_flat}(b) or equivalently in Figure~\ref{fig:I_Q_3D}(b). The noise is considered to be an independent additive Gaussian random variable. The ratio between the signal and the noise power levels is set by the \CNZERO value, a classic figure of merit for quality of reception in \gls{GNSS} receivers. In addition may come a multi-path distortion parameterized by its deviation in amplitude, time, frequency and phase with respect to the \gls{LOS} signal. The model of the multi-path distortion is hence the same as for the \gls{LOS} signal drawn in Figures~\ref{fig:I_Q_flat}(b) or~\ref{fig:I_Q_3D}(b) but attenuated in amplitude, shifted in time and frequency, and rotated in phase between $I$ and $Q$. The probability density functions and ranges of the parameters are detailed in~\cite{Gonzalez2024}. Figures~\ref{fig:I_Q_flat}(c) and~\ref{fig:I_Q_3D}(c) represent a multi-path signal characterized by a specific phase rotation of 90° which causes a swap between picture $I$ and picture $Q$ as they are orthogonal representations of the signal. The aggregation of the \gls{LOS} signal and of the multi-path interference is pictured in Figures~\ref{fig:I_Q_flat}(d) and~\ref{fig:I_Q_3D}(d). Finally, Figures~\ref{fig:I_Q_flat}(a) and~\ref{fig:I_Q_3D}(a) are an example of a pair of full composite images produced by the generator. It is the addition of noise to the aggregation of the \gls{LOS} signal and of the multi-path interference represented in Figures~\ref{fig:I_Q_flat}(c) and~\ref{fig:I_Q_3D}(c).

Each pair of images is labelled with a \emph{0} if it is distortion-free or with a \emph{1} if it is polluted by a multi-path signal.

\begin{figure}
\newlength{\mywiqf}
\setlength{\mywiqf}{0.68\textwidth}
\setlength{\belowcaptionskip}{0.6\baselineskip}
\centering
\subfloat[\centering Full image]{{
\includegraphics[trim={300 500 180 350}, clip, width=\the\mywiqf]{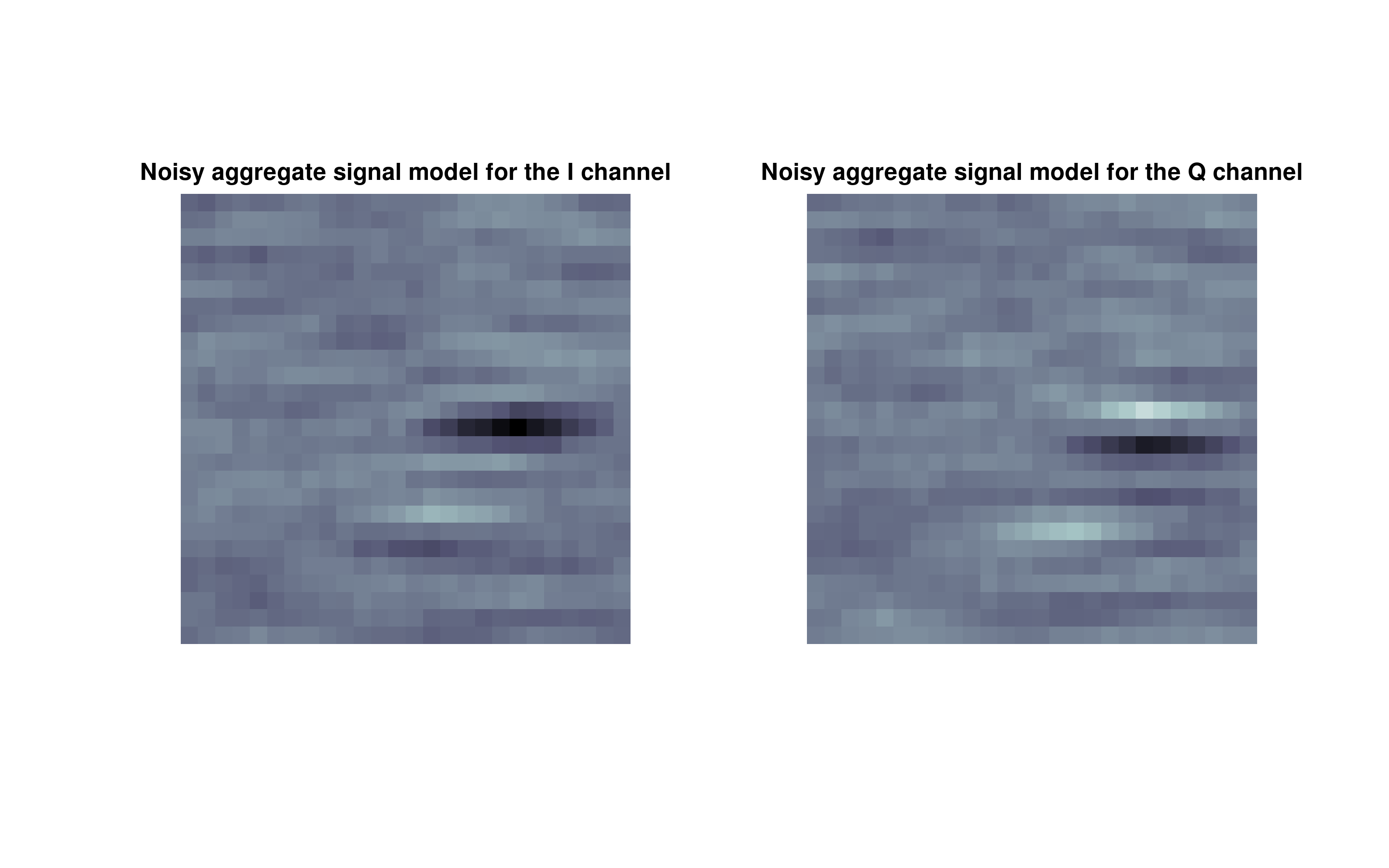} }}
\\
\subfloat[\centering Direct signal model]{{
\includegraphics[trim={300 500 180 350}, clip, width=\the\mywiqf]{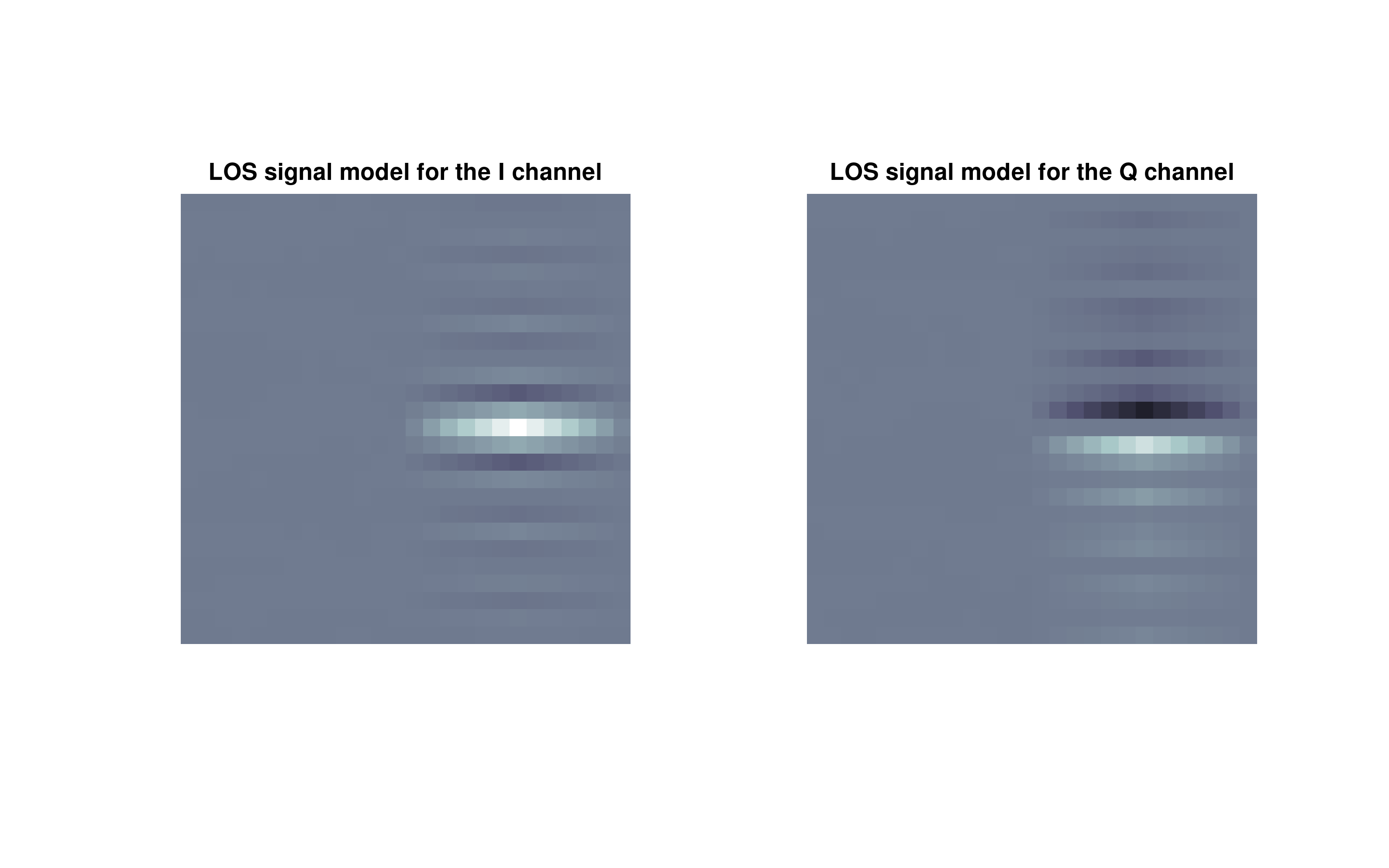} }}
\\
\subfloat[\centering Multi-path signal model]{{
\includegraphics[trim={300 500 180 350}, clip, width=\the\mywiqf]{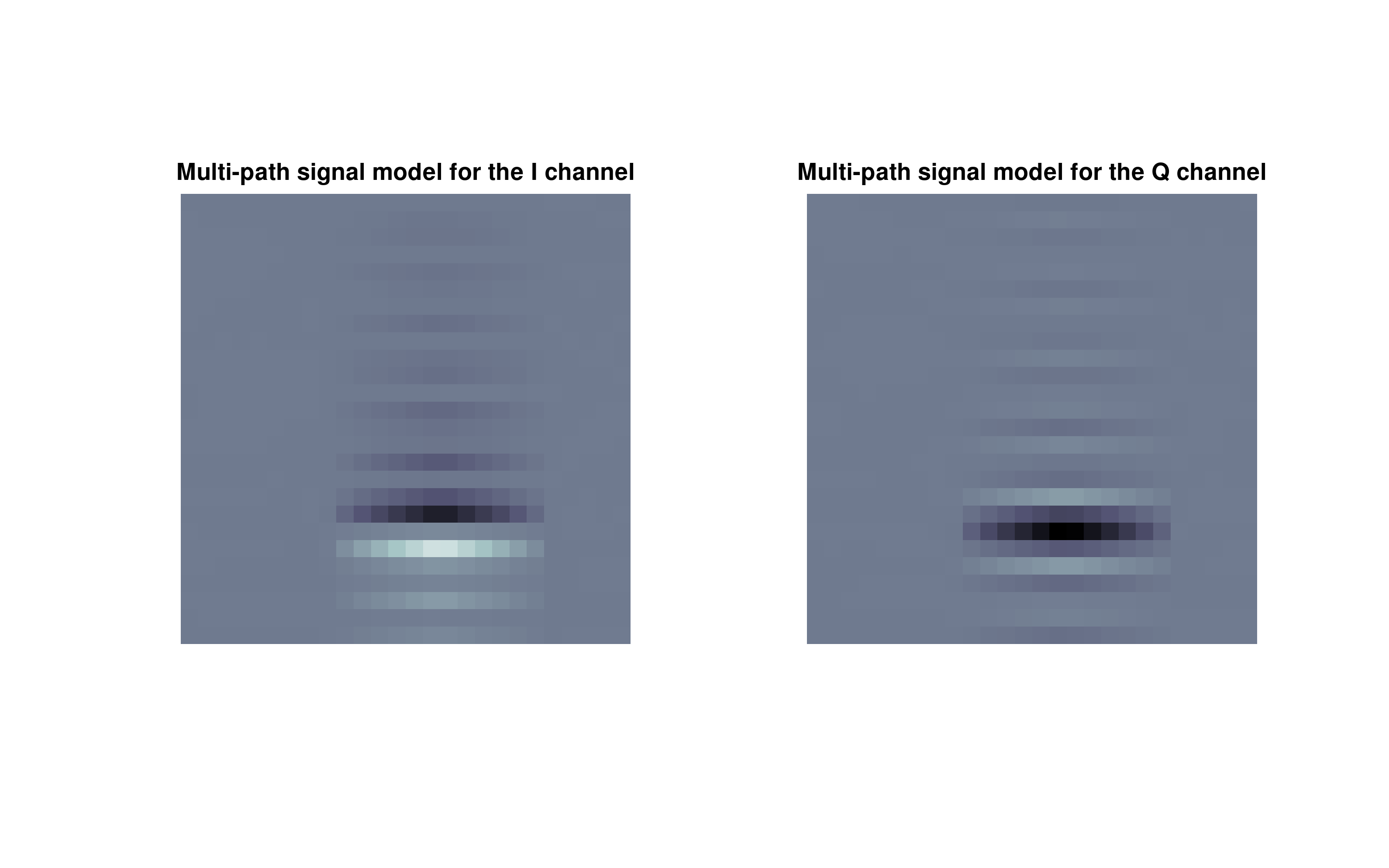} }}
\\
\subfloat[\centering Aggregate signal model]{{
\includegraphics[trim={300 500 180 350}, clip, width=\the\mywiqf]{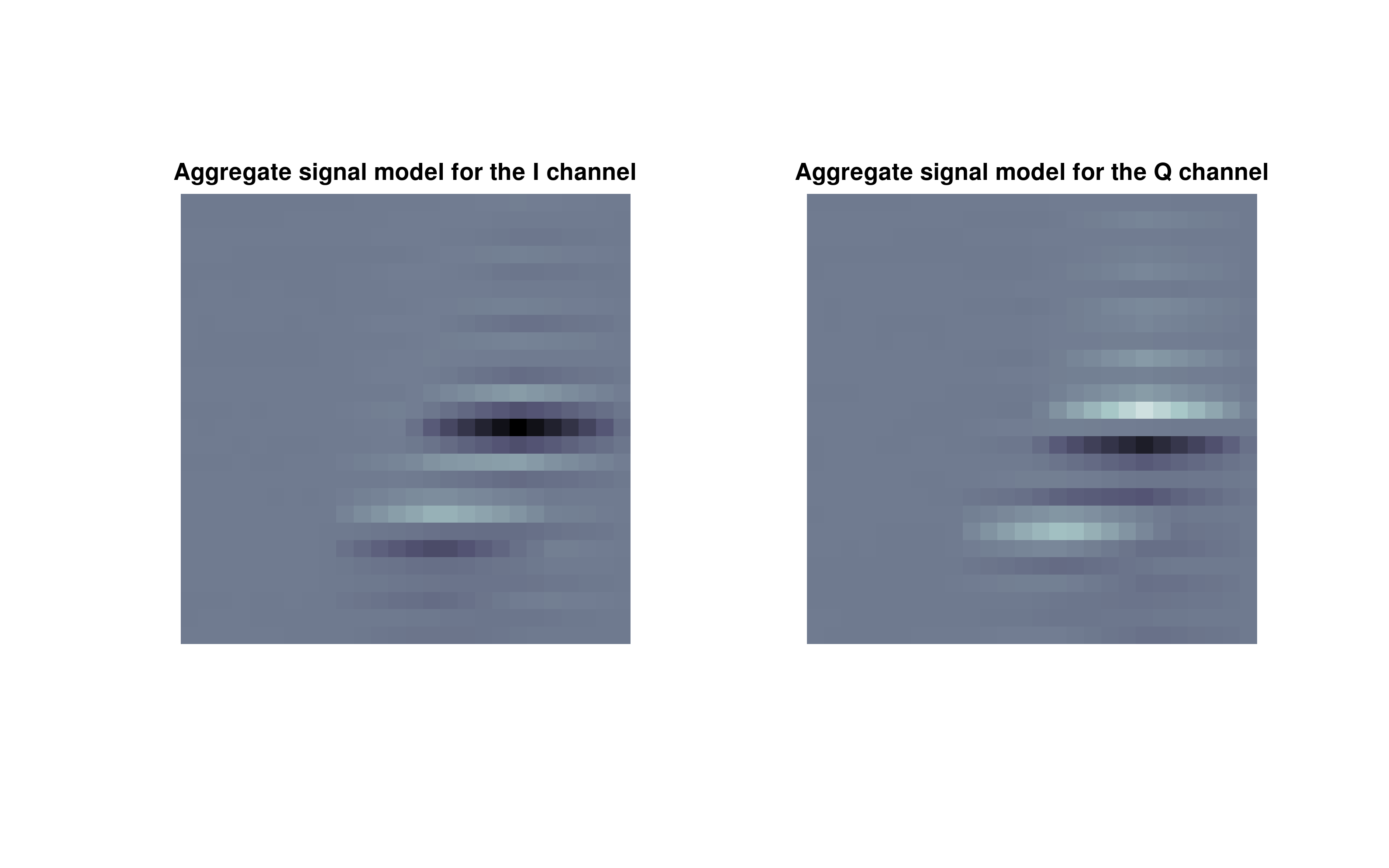} }}
\caption{Examples of pairs of $I$ and $Q$ images from the data generator.}
\label{fig:I_Q_flat}
\end{figure}

\subsection{Wasserstein distance metric for multi-path detection}
\label{section_Wasserstein}
The input data for multi-path detection are composed of \gls{2-D} phased images spread on two channels ($I$ and $Q$) and could be seen as images in the complex space, meaning that each pixel of an image has a complex value ($I$ corresponding to the real part and $Q$ the imaginary part). As recalled below, an efficient metric to compare two images is to consider the Wasserstein distance that computes the minimum energy cost to map one image seen as a probability distribution to the other image.

However, to the best of our knowledge, there is no computational procedure to compute this metric when images are seen as complex measures. For this reason, we propose to compute the Wasserstein distance between two pairs $(I,Q)$ and $(I',Q')$ of images by independently computing the optimal transportation cost between the pair $(I,I')$ first and then $(Q,Q')$. The sum of the two optimal transportation costs will then be considered as the distance between the two pairs $(I,Q)$ and $(I',Q')$ of images. \\

\noindent\textbf{The Wasserstein distance as an optimisation problem}

The underlying theoretical framework for computing the Wasserstein distance between two images $a$ and $b$ is the Monge–Kantorovitch optimal transport problem~\cite{Peyre2019,Friesecke2025}. Consider $\{a_i\}$ for $i \in \{1, \dots, n\}$ and $\{b_j\}$ for $j \in \{1, \dots, m\}$ the pixel values of $a$ and $b$ respectively. Assume also that the images are offset and scaled so that $a_i \ge 0 \text{ and } \sum_{i}{a_i} = 1$ and $b_j \ge 0 \text{ and } \sum_{j}{b_j} = 1$ so that they can be seen as discrete probability distributions. In that case, the Wasserstein distance is the minimum energy required to transport image $a$ to image $b$ and can be formulated as the optimal objective value of the following optimization problem:
\begin{eqnarray*}
\dsp \min_{P\in [0, 1]^{n \times m}} &\dsp\sum_{i=1}^{n}\sum_{j=1}^{m} P_{ij}C_{ij}\\
\mbox{with respect to}&\\
&\dsp\sum_{j=1}^{m} P_{ij}=a_i \qquad \forall i=1\ldots,n, \\
&\dsp\sum_{i=1}^{n} P_{ij}=b_j \qquad \forall j=1,\ldots,m.
\end{eqnarray*}

\noindent where $P$ is the unknown mass transportation matrix and $C$ the given matrix of Euclidean distances between pixels.\\
The first set of constraints for each $i$ corresponds to the conservation of the mass of image $a$ during transport while the second set of constraints for each $j$ represents the conservation of the mass received in image $b$. The above optimization problem is a linear programming problem and can be efficiently solved even for large dimensional images. This computation will be carried out for each pair of image $(I,I')$ or $(Q,Q')$ in the GNSS application.\\

\noindent Note on the computation of the pixel distance matrix $C$: the Euclidean distance between pixels involved in the above optimization problem is not sample dependent and can be computed once and applied during the whole iterative calculation process of the Wasserstein distance. To do so, images are first flattened using a flattening function $flatten: \RR^n \times \RR^n \rightarrow \RR^{n^2}: (k,l)\rightarrow l+(k-1)n$, then the Euclidean distance between the pixels $(p,q)$ and $(k,l)$ of two images is given by $C_{ij}= \sqrt{(p - k )^2 + (q - l)^2}$ where $i = flatten(p,q)$ and $j = flatten(k,l)$.

\section{Experiments}
\label{section_experiments}
In this section, our \gls{SSL} experimental framework is detailed. Firstly, the \gls{CNN} architecture is displayed. Then, the data sets are exposed in depth. Lastly, the methodology of the experiments is explained.

\subsection{\gls{CNN} architecture}
\begin{figure}
\begin{center}
\includegraphics[scale=0.23]{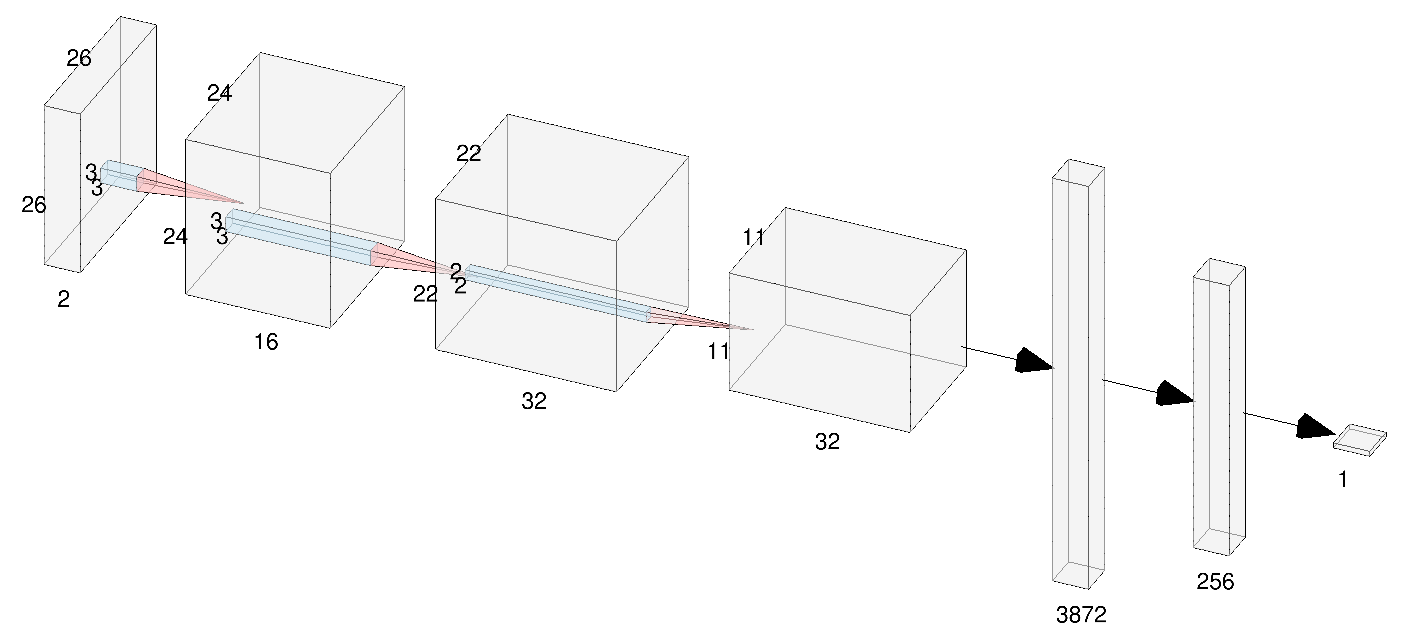}
\caption{Architecture of the \gls{CNN} used in this work~\cite{LeNail2019}.}
\label{figure:CNN_architecture}
\end{center}
\end{figure}

The \gls{CNN} architecture implemented in this work is depicted in Figure~\ref{figure:CNN_architecture}. It is part of the \gls{VGG} network class \cite{Simonyan2015}. This kind of neural network has been obsoleted for long by other structures such as ResNet~\cite{he2015deep}, Inception~\cite{szegedy2015rethinking} or DenseNet \cite{Huang2018}. However, our approach in this study is to emphasise the methodological aspects of the \gls{SSL} technique rather than detail a fully optimised neural architecture framework. For this reason, we have chosen to use a baseline neural network architecture (the \gls{VGG} architecture) that is commonly accepted as efficient in a wide range of applications on images. Using alternative architectures the proposed method would be exactly identical as the one detailed in this article. The model performance could even achieve better results as an optimal neural architecture could be found for the application at hand. However, this would not highlight the role of \gls{SSL} but rather the optimisation of the architecture. The choice of architecture is therefore secondary and this is why in the sequel of the article we will only consider a \gls{VGG}-like neural network.

The input tensor made of a pair of 26x26 dimensional $I$ and $Q$ images is first analysed by a set of 16 filters of dimension 3x3. Next, an other set of 32 filters of dimension 3x3 is applied, followed by a 2x2 dimensional max pooling layer which completes the convolutional part of the network. The data are then flattened to feed a 3872 neurons linear layer. A layer of 256 neurons with a \gls{RELU} activation function takes over to feed the final output neuron which makes use of a sigmoïd activation function.

The \gls{CNN} code has been written in the Python language using the PyTorch library \cite{Paszke2019}. More specifically, the key function \texttt{ot.lp.emd} of the \gls{POT} library \cite{Flamary2024} has been used to calculate the Wassertein distance between pair of images. Again, the \gls{CNN} code \cite{Blais2025b} is freely available under the terms of the GNU \gls{GPL} version 3 license.

\subsection{Data sets composition}
\label{section_data_sets}
The data sets used for training and validation were elaborated in accordance with the \gls{GNSS} context as well as with our \gls{SSL} assessment objective.
At the highest level, three \CNZERO ratios have been considered to address realistic operating conditions of a \gls{GNSS} receiver: 
\begin{itemize}
\item 37 \dBHz, corresponding to rather poor receiving conditions,
\item 40 \dBHz, which could be interpreted as a mid-range value,
\item 43 \dBHz, reflecting fairly good receiving conditions.
\end{itemize}

Then, for each \CNZERO ratio the distribution between labelled and unlabelled samples was made has follows for training:
\begin{itemize}
\item The total number of samples was set to $200$. This is the sum of the number of labelled samples $N_\text{SUP}$ and of the number of unlabelled samples $N_\text{UNSUP}$. It will be shown in section~\ref{section_fully_supervised} that at $N_\text{SUP} = 200$ the classification accuracy is not growing significantly anymore and no appreciable learning gain is expected for greater $N_\text{SUP}$.
\item The values of $N_\text{SUP}$ have been staggered in the set $\{25, 40, 50, 60, 75\}$. Again, it will be emphasized in section~\ref{section_fully_supervised} that below $N_\text{SUP} = 25$ training is unreliable while above $N_\text{SUP} = 75$ we are in situations where the majority of samples are labelled and these use cases are out of the scope of this study.
\end{itemize}
For the validation step, the number of samples, labelled by definition, was systematically set to $100$.

Finally, the training and validation data sets were balanced with respect to the number of samples disturbed by a multi-path or not. The parameters of the multi-path, amplitude, time, frequency and phase, are not detailed here but they are presented in detail in~\cite{Gonzalez2024}. They were also set to reflect realistic receiving conditions encountered by a \gls{GNSS} receiver.

\subsection{Methodology}
In order to establish the performance of \gls{SSL}, for each measurement point the following procedure was applied for training and validation of the network:
\begin{itemize}
\item 299 runs were executed. A run is one learning experience, starting from random initialisation of the weights of the network to the convergence of the loss function close to zero.
\item 55 epochs have been carried out in each run. An epoch is defined as the complete scan of the training data set. This value of 55 was high enough to ensure the convergence of the loss function,
\item At the end of each epoch a validation step was carried out, calculating the accuracy over the full validation data set.
\end{itemize}

More specifically about training, it is also worth noting that $\mathcal{L}_S$ is defined as the \gls{BCE} and that the \texttt{ADAM} optimizer~\cite{kingma2017} was used with a batch size of 50.

Concerning the validation step, the figure of merit we selected to report the validation performance is the median of the maximum accuracy. More precisely, for a given measurement point, in each run the maximum value of the accuracy is read over all epochs, then the median value is calculated over the 299 maxima. The raw maximum accuracy has not been retained as we have considered it would have been an over-optimistic metric.

At the highest level, a grid search was conducted to find the best pairs of parameters $(\lambda, \sigma)$, for each \CNZERO ratio and each $N_\text{SUP}$ listed herebefore. The grid was built with $\lambda \in \{1, 10, 100, 1000\}$ and $\sigma \in \{0.01, 0.05, 0.1, 0.2, 0.5, 1.0, 2.0, 10.0\}$.

The implementation of the test bench used to build the data sets described in section~\ref{section_data_sets} as well as to apply the experimental methodology detailed in this section is again freely available in the same repository which hosts the code of the \gls{CNN}~\cite{Blais2025b}.

\section{Results and discussion}
\label{section_results}

This section presents and discusses the outcomes of our experiments. To set a reference for \gls{SSL}, the performance of fully supervised learning is first established. Then, the results for \gls{SSL} are exposed in comparison to the reference. The results for the pairs of parameters $(\lambda, \sigma)$ providing the largest gain over fully supervised learning are also detailed.

\subsection{Performance of the fully supervised \gls{CNN}}
\label{section_fully_supervised}
\begin{figure}
\centering
\includegraphics[trim=130 30 130 50, clip, width=1.0\textwidth]{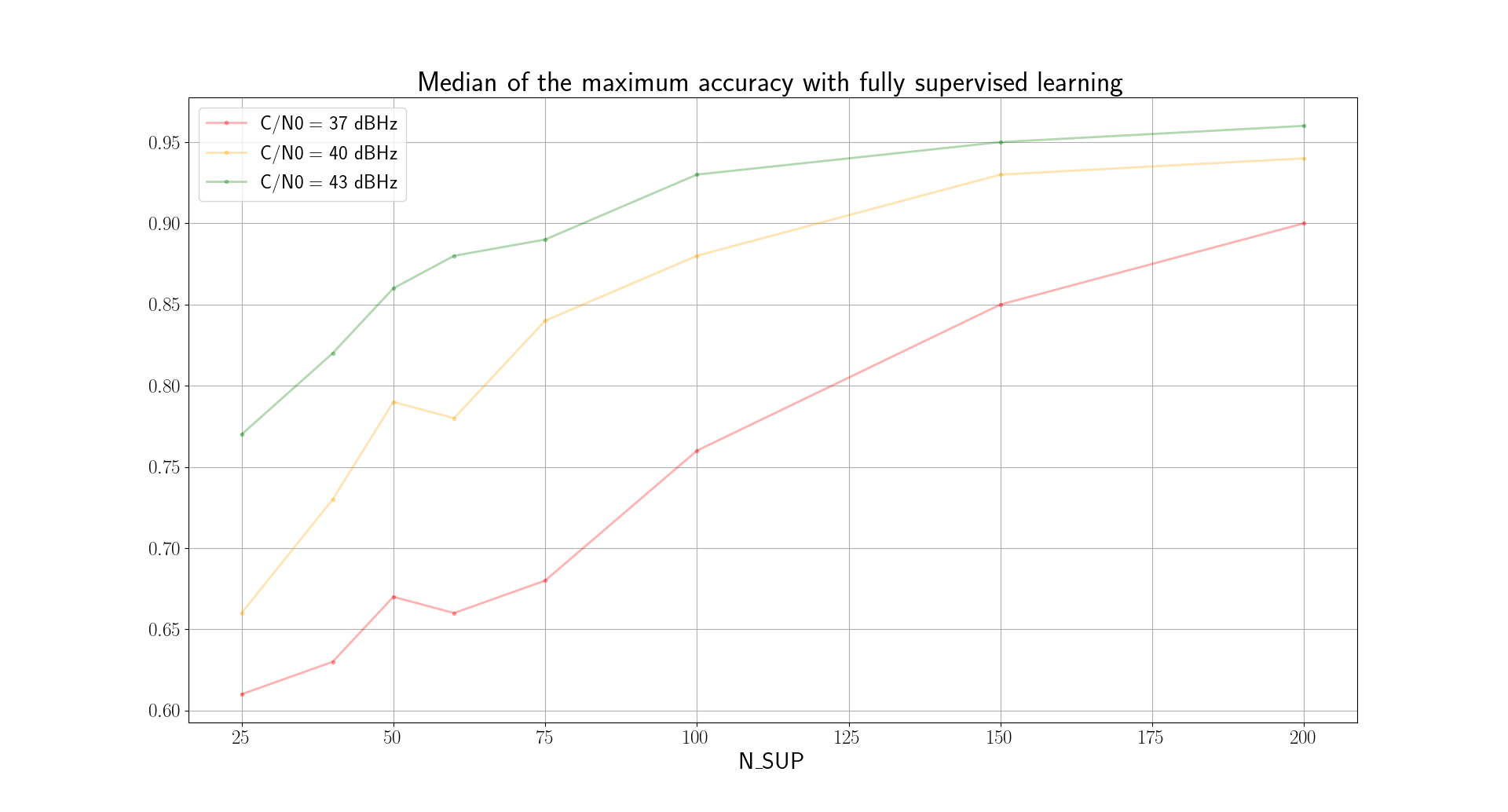}
\caption{Median maximum accuracy for the fully supervised reference cases.}
\label{figure:ref_cases}
\end{figure}

The median maximum accuracy achieved by the fully supervised \gls{CNN} with respect to the number of training (labelled) samples $N_\text{SUP}$ is displayed in Figure~\ref{figure:ref_cases}. The range of $N_\text{SUP}$ is limited to $[25, 200]$ as below 25 learning turned out to be unreliable and above 200 the performance has reached a logarithmic growth region.

The effect of the \CNZERO ratio on the network classification efficiency is clearly visible and coherent with the logic that for the same number of training samples the higher the signal to noise ratio is the higher the performance shall be. Besides, the three curves show two distinct areas. The first is a linearly increasing accuracy region for low $N_\text{SUP}$, which extends up to about 60 for \CNZERO = 43 \dBHz, 75 for \CNZERO = 40 \dBHz and 150 for \CNZERO = 37 \dBHz. One can observe that the lower the \CNZERO is, the more extended this region is. In this region the effect of a small increase in the number of labelled samples greatly improves the results and even more so when the \CNZERO ratio increases. The second region, for larger $N_\text{SUP}$, demonstrates a reduction in the performance improvement with $N_\text{SUP}$. Indeed, the behaviour is plainly logarithmic for \CNZERO = 43 \dBHz and \CNZERO = 40 \dBHz.

\begin{figure}
\edef\trimopt{trim = 30 0 30 0, clip, width=0.48\textwidth}
\setlength{\belowcaptionskip}{0.4\baselineskip}
\centering
\subfloat[\centering C/N0 = 37 dBHz.]{{\expandafter\includegraphics\expandafter[\trimopt]{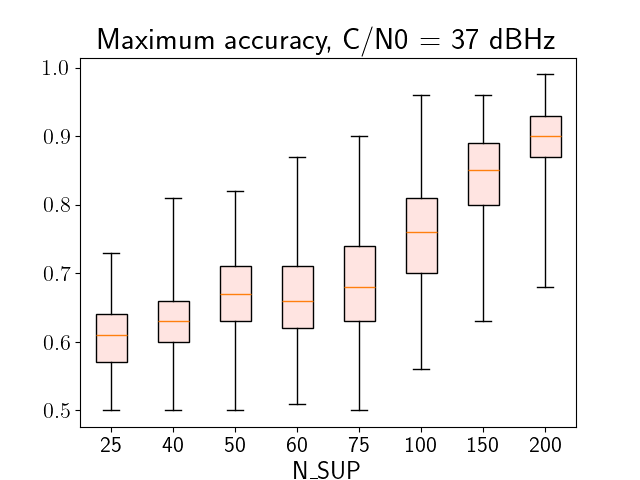} }}
\qquad
\subfloat[\centering C/N0 = 40 dBHz.]{{\expandafter\includegraphics\expandafter[\trimopt]{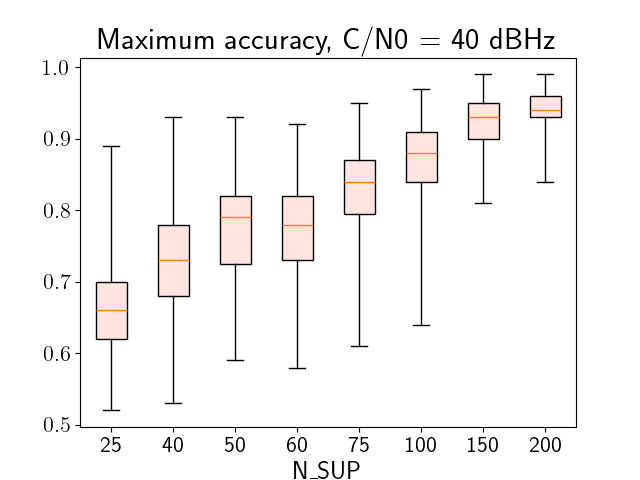} }}
\subfloat[\centering C/N0 = 43 dBHz.]{{\expandafter\includegraphics\expandafter[\trimopt]{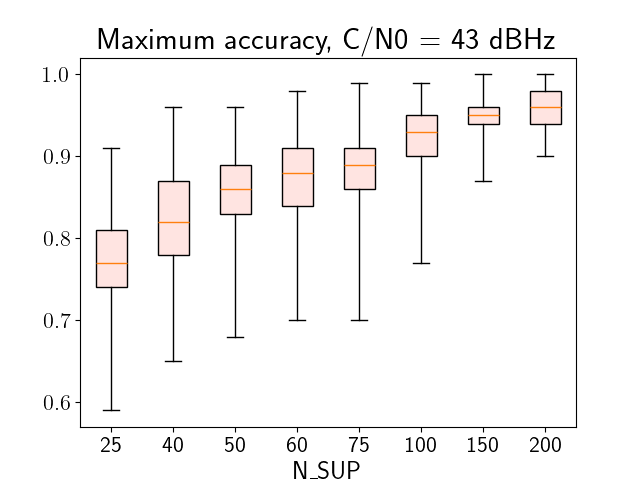} }}
\caption{Quartiles of the maximum accuracy for the fully supervised reference cases.}
\label{figure:ref_cases_detail}
\end{figure}

Figure~\ref{figure:ref_cases_detail} presents the quartiles for the reference results of Figure~\ref{figure:ref_cases}. The size of the quartiles does not exhibit the same trends as the median value. There is no significant shrinking of the quartiles with an increase in $N_\text{SUP}$ nor in \CNZERO, except for few cases when $N_\text{SUP} \ge 150$ and \CNZERO $\ge$ 40 \dBHz.

In light of Figure~\ref{figure:ref_cases}, the rest of the study has focused on $N_\text{SUP} \le 75$ as above 75 the performance of the fully supervised \gls{CNN} is already so high that the margins for improvement has seemed relatively weak.

\subsection{Performance of \gls{SSL} and comparison with fully supervised learning}

\begin{figure}
\centering
\includegraphics[trim=130 30 100 50, clip, width=1.0\textwidth]{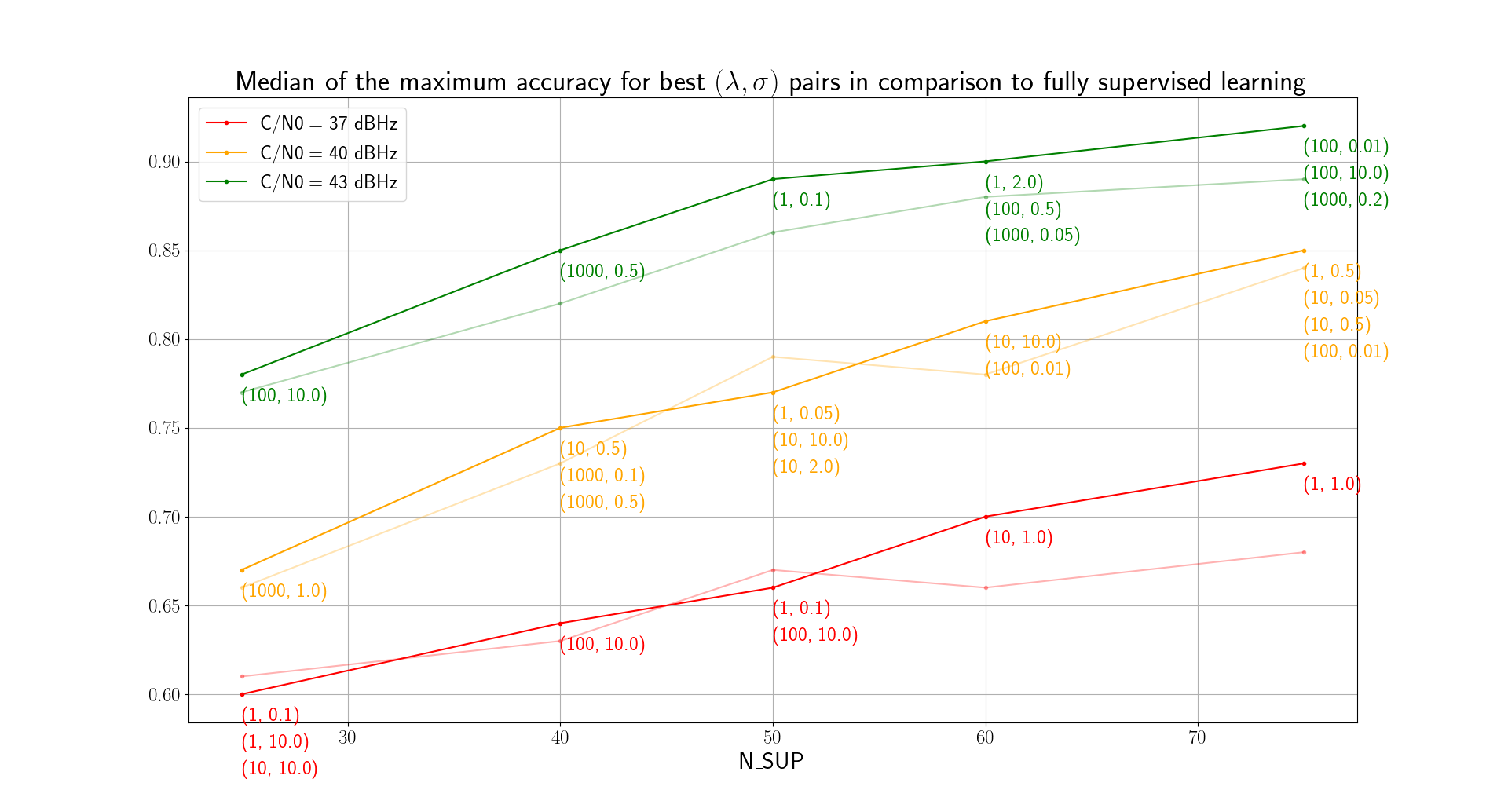}
\caption{Median maximum accuracy for the \gls{SSL} best pairs of parameters $(\lambda, \sigma)$ and comparison with fully supervised learning. Bold curves represent the \gls{SSL} results and the thin curves are for the fully supervised learning from Figure~\ref{figure:ref_cases}.}
\label{figure:ssl_cases}
\end{figure}

Figure~\ref{figure:ssl_cases} shows the best results obtained with the \gls{SSL} \gls{CNN} (bold curves) and recalls the performance of the fully supervised \gls{CNN} (thin curves) as well. The pairs of parameters which have enabled the \gls{SSL} network to achieve each particular score are also printed under the corresponding point of the curve.

Except for $N_\text{SUP} = 25$ in combination with \CNZERO = 37 \dBHz and $N_\text{SUP} = 50$ in combination with \CNZERO = 40 \dBHz and 37 \dBHz, there always exists at least one pair of parameters $(\lambda, \sigma)$ which allows \gls{SSL} to outperform fully supervised learning. The gain in median maximum accuracy can reach 5 \% in the specific case of $N_\text{SUP} = 75$ for \CNZERO = 37 \dBHz and is never less than 1 \% apart from the cases listed before. More precisely, on a total of 15 measurement points, 1 shows a loss of 2 \% in performance, 2 a loss of 1 \%, 4 a gain of 1 \%, 2 a gain of 2 \%, 4 a gain of 3 \%, 1 a gain of 4 \% and 1 a gain of 5 \%. The balance is clearly in favour of \gls{SSL}. Indeed, Figure~\ref{figure:ssl_cases} validates the relevance of our approach by showing that the introduction of unlabelled data improves the general learning performance of our model.

However, one can notice that no clear rule could be established to predetermine the pair(s) of parameters $(\lambda, \sigma)$ which would bring the best result in any given conditions set by $N_\text{SUP}$ and \CNZERO.

\begin{figure}
\centering
\includegraphics[trim=30 0 30 0, clip, width=0.48\textwidth]{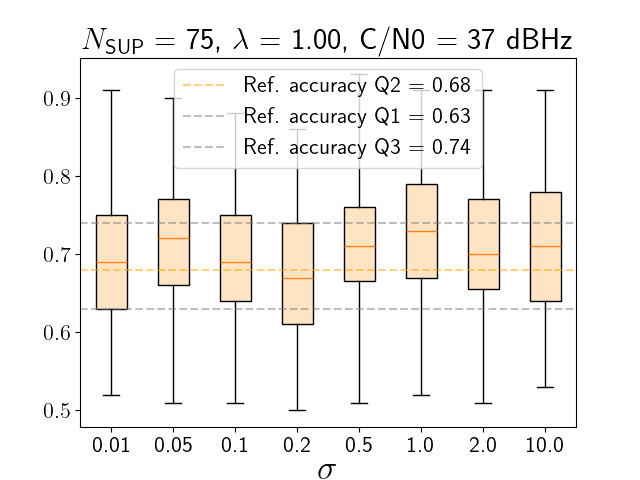}
\caption{Quartiles of the maximum accuracy for \CNZERO = 37 \dBHz, $N_\text{SUP} = 75$.}
\label{fig:best_37}
\end{figure}

\begin{figure}
\edef\trimopt{trim = 30 0 30 0, clip, width=0.48\textwidth}
\centering
\subfloat[\centering $\lambda = 10$.]{{\expandafter\includegraphics\expandafter[\trimopt]{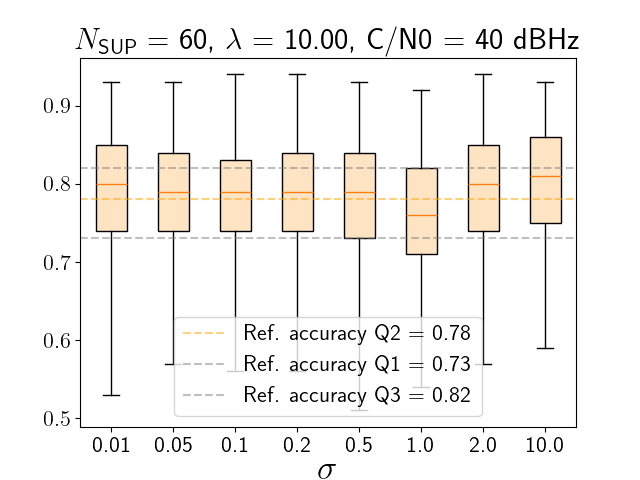} }}
\subfloat[\centering $\lambda = 100$.]{{
\expandafter\includegraphics\expandafter[\trimopt]{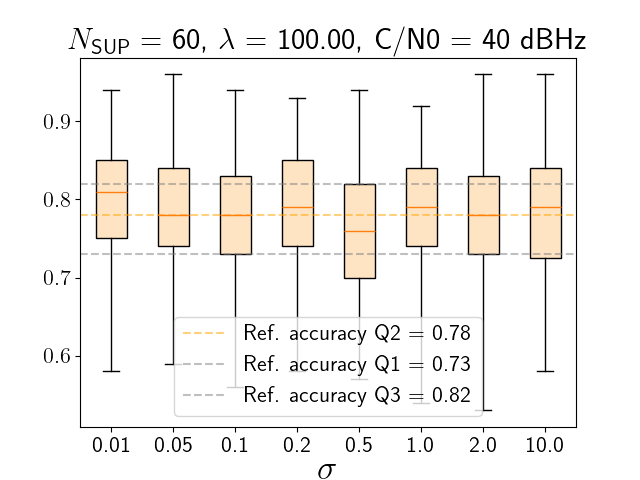} }}
\caption{Quartiles of the maximum accuracy for \CNZERO = 40 \dBHz, $N_\text{SUP} = 60$.}
\label{fig:best_40}
\end{figure}

\begin{figure}
\edef\trimopt{trim = 30 0 30 0, clip, width=0.48\textwidth}
\setlength{\belowcaptionskip}{0.4\baselineskip}
\centering
\subfloat[\centering $N_\text{SUP} = 40$ and $\lambda = 1000$.]{{
\expandafter\includegraphics\expandafter[\trimopt]{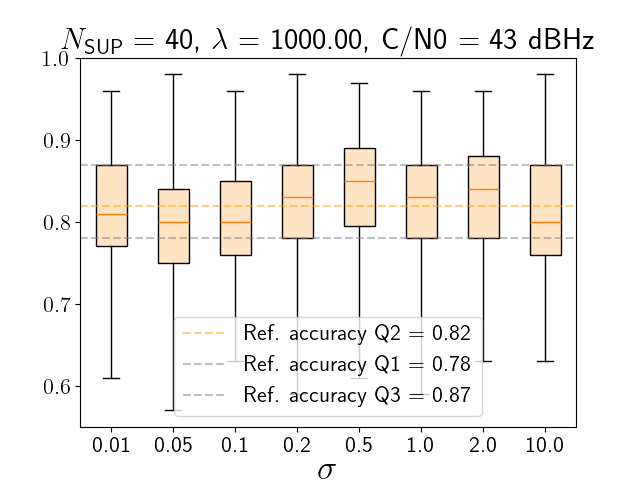} }}
\subfloat[\centering $N_\text{SUP} = 50$ and $\lambda = 1$.]{{
\expandafter\includegraphics\expandafter[\trimopt]{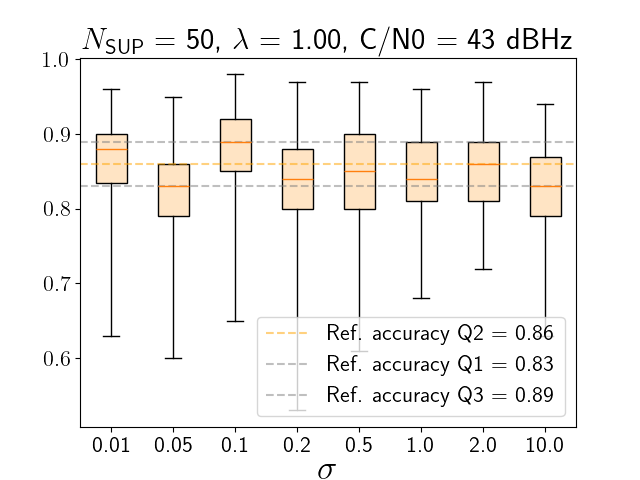} }}
\qquad
\subfloat[\centering $N_\text{SUP} = 75$ and $\lambda = 100$.]{{
\expandafter\includegraphics\expandafter[\trimopt]{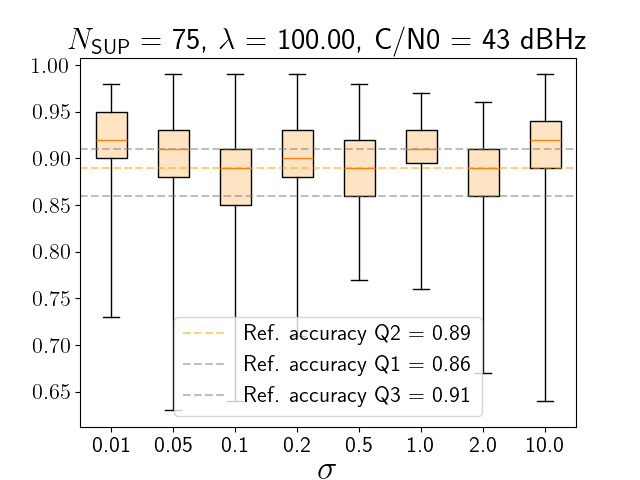} }}
\subfloat[\centering $N_\text{SUP} = 75$ and $\lambda = 1000$.]{{
\expandafter\includegraphics\expandafter[\trimopt]{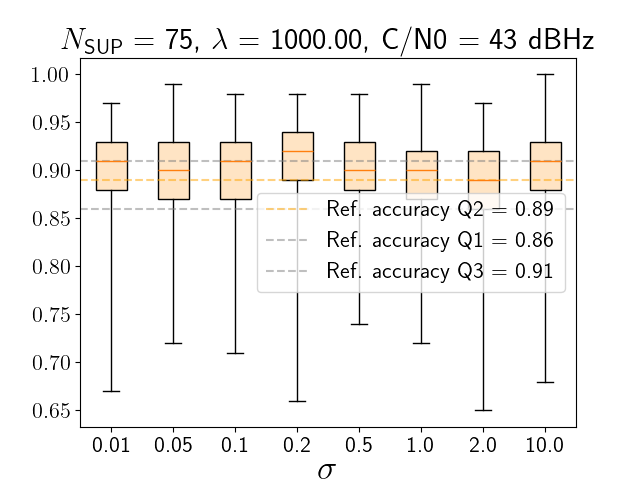} }}
\caption{Quartiles of the maximum accuracy for \CNZERO = 43 \dBHz.}
\label{fig:best_43}
\end{figure}

Figures~\ref{fig:best_37}, ~\ref{fig:best_40} and~\ref{fig:best_43} detail the quartiles for the best results of Figure~\ref{figure:ssl_cases}. For each plot the \CNZERO ratio, $N_\text{SUP}$ and $\lambda$ are set and the quartiles of the maximum accuracy are plotted for the different values of $\sigma$. More specifically, these figures depict the following cases,
\begin{itemize}
\item Figure~\ref{fig:best_37} for 37 \dBHz: a gain of 5 \% for $N_\text{SUP} = 75$ with $(\lambda, \sigma)$ = (1, 1.0),
\item Figure~\ref{fig:best_40} for 40 \dBHz: a gain of 3 \% for $N_\text{SUP} = 60$ in
\begin{itemize}
    \item Sub-figure (a) with $(\lambda, \sigma)$ = (10, 10)
    \item Sub-figure (b) with $(\lambda, \sigma)$ = (100, 0.01).
\end{itemize}
\item Figure~\ref{fig:best_43} for 43 \dBHz: a gain of 3 \% in
\begin{itemize}
    \item Sub-figure (a) for $N_\text{SUP} = 40$ with $(\lambda, \sigma)$ = (1000, 0.5),
    \item Sub-figure (b) for $N_\text{SUP} = 50$ with $(\lambda, \sigma)$ = (1, 0.1),
    \item Sub-figure (c) for $N_\text{SUP} = 75$ with $(\lambda, \sigma)$ = (100, 0.01) and (100, 10.0),
    \item Sub-figure (d) for $N_\text{SUP} = 75$ also with $(\lambda, \sigma)$ = (1000, 0.2).
\end{itemize}
\end{itemize}

The quartile plots for the other values of $N_\text{SUP}$, \CNZERO and $\lambda$ are available in the code repository already mentioned.

In each plot the quartiles of the reference case corresponding to the values of \CNZERO and $N_\text{SUP}$ are indicated by horizontal grey dashed lines.

Exactly as discussed in section~\ref{section_fully_supervised} no significant variation in the quartile size arises from these plots as a function of the number of labelled samples $N_\text{SUP}$ or depending on the \CNZERO ratio. There is one exception, though, when $N_\text{SUP} = 25$ for the three values of \CNZERO the performance of \gls{SSL} is nearly every time significantly lower than for fully supervised learning. Figure~\ref{figure:25_cases} which can be found in Appendix~\ref{appendix_N_SUP_25} illustrates this observation. It could indicate that $N_\text{SUP} = 25$ labelled samples are too scarce for the network to learn accurately the manifold of the data. On another note, the comparison with the quartiles of the reference cases shows that they are similar in size indicating that the introduction of unlabelled samples, even in large number as for $N_\text{SUP} = 40$ with then $N_\text{UNSUP} = 160$, does not degrade the reliability of the classification statistics.

\section{Conclusions}
\label{section_conclusions}

In this research work, we have proposed a semi-supervised classification method for images in situations where labelled data are scarce. The use of the Wasserstein distance as a similarity metric between image samples has enabled the graph-based transductive label propagation in the proposed \gls{CNN} architecture. Furthermore, the study has demonstrated its efficiency on a real-life \gls{GNSS} application. The results show that for various receiving conditions a gain in accuracy of up to 5 \% can be achieved. This emphasizes the existence of some form of regularity in the \gls{GNSS} data which can be exploited. We believe that it is also the case in many practical applications, accuracy results can be improved by introducing unlabelled samples in the learning process. This advocates also for the use of such techniques to reduce the generally expensive labelling procedure of \emph{ground truth} data. However, so far no clear rule has been found to select the optimum values for the hyperparameters involved in the method and further investigation in this direction should be conducted. \\

\bibliographystyle{apalike}
\bibliography{refs}

\appendix

\section{I and Q images represented in \gls{3-D}}
\label{appendix_I_Q_3D}

\begin{figure}
\newlength{\mywiq}
\setlength{\mywiq}{0.45\textwidth}
\setlength{\belowcaptionskip}{0.6\baselineskip}
\centering
\subfloat[\centering Full image]{{
\includegraphics[trim={300 200 180 50}, clip, width=\the\mywiq]{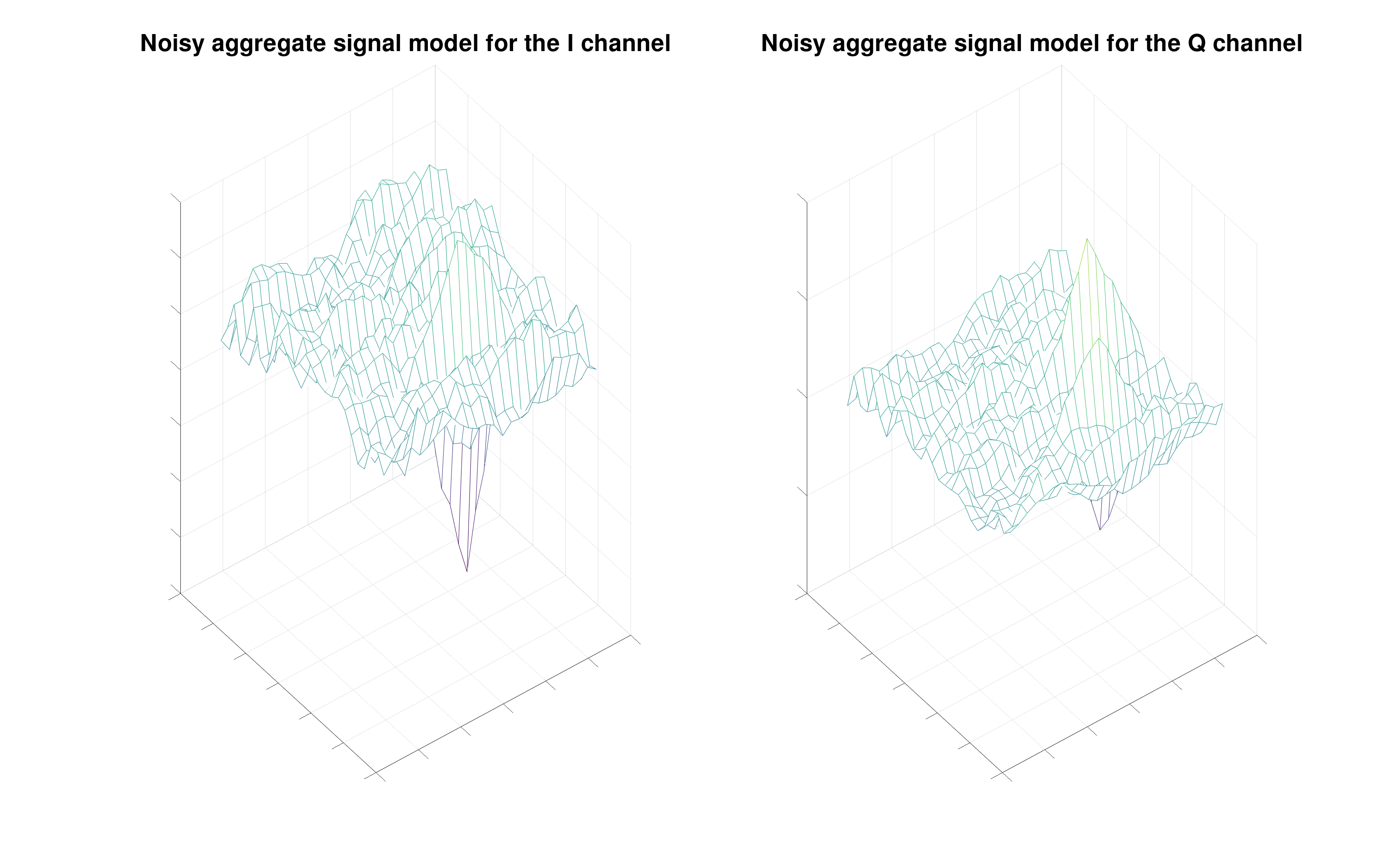} }}
\\
\subfloat[\centering Direct signal model]{{
\includegraphics[trim={300 200 180 50}, clip, width=\the\mywiq]{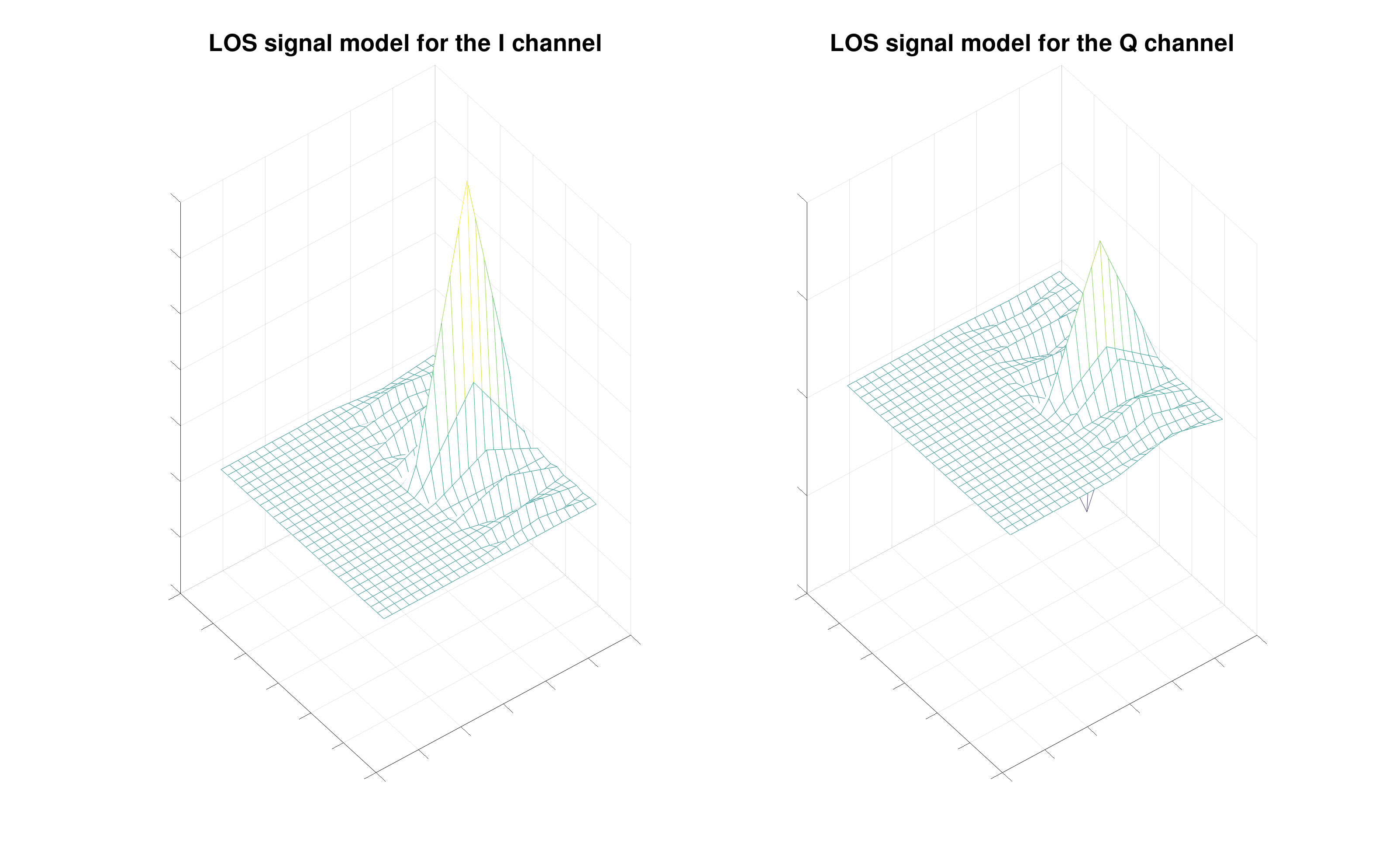} }}
\\
\subfloat[\centering Multi-path signal model]{{
\includegraphics[trim={300 200 180 50}, clip, width=\the\mywiq]{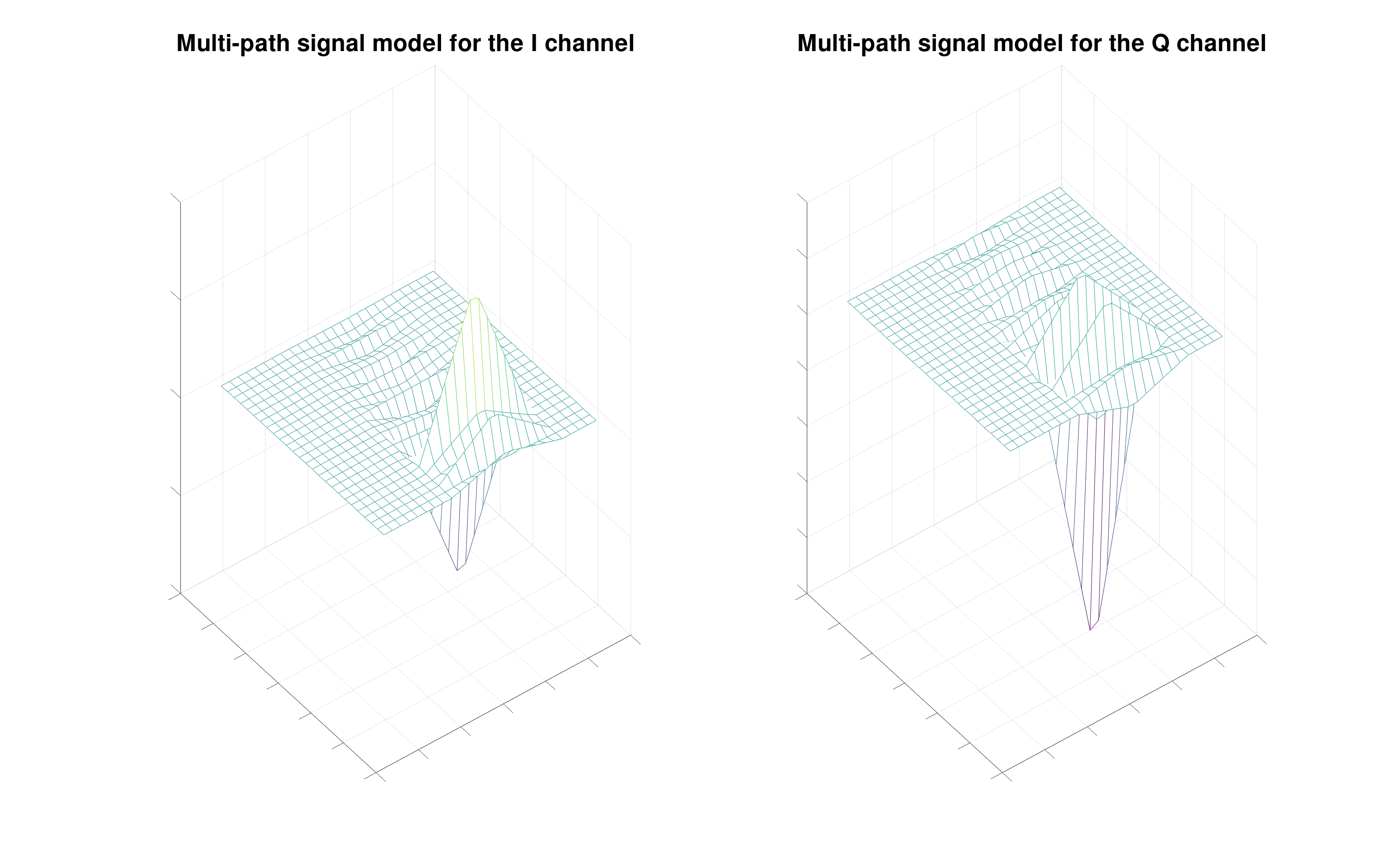} }}
\\
\subfloat[\centering Aggregate signal model]{{
\includegraphics[trim={300 200 180 50}, clip, width=\the\mywiq]{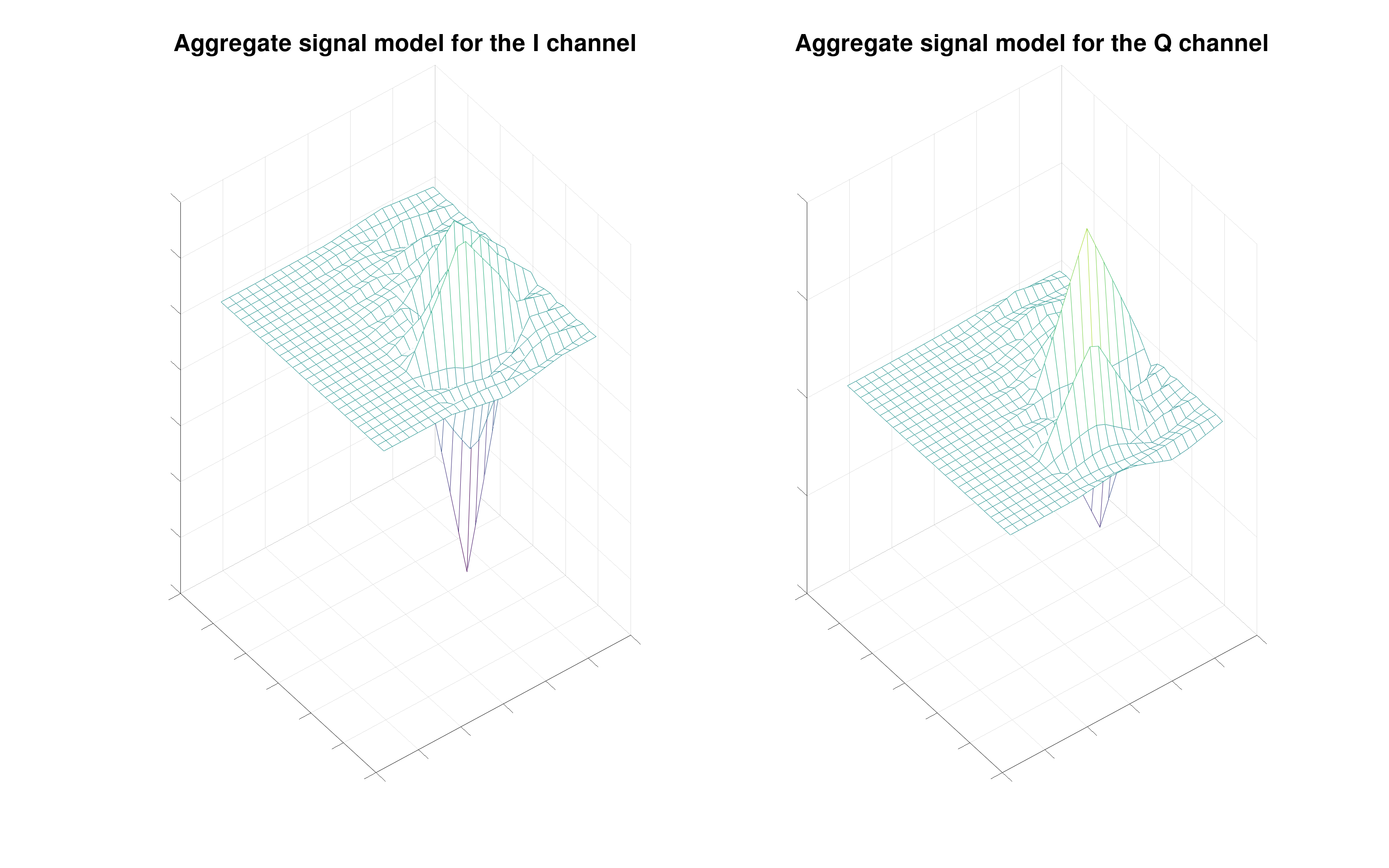} }}
\caption{The same pairs of $I$ and $Q$ images as in Figure~\ref{fig:I_Q_flat} represented in \gls{3-D}.}
\label{fig:I_Q_3D}
\end{figure}

\section{Quartile plots for N\texorpdfstring{\textsubscript{SUP}}{N SUP} = 25}
\label{appendix_N_SUP_25}

\begin{figure}
\edef\trimopt{trim = 30 0 30 0, clip, width=0.48\textwidth}
\setlength{\belowcaptionskip}{0.4\baselineskip}
\centering
\subfloat[\centering \CNZERO = 37 \dBHz.]{{\expandafter\includegraphics\expandafter[\trimopt]{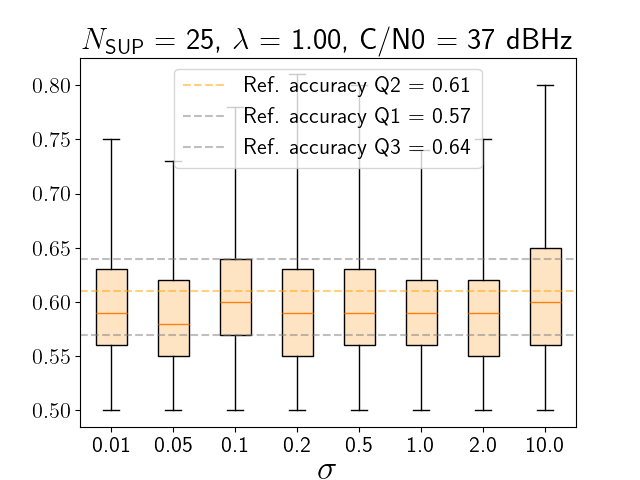} }}
\qquad
\subfloat[\centering \CNZERO = 40 \dBHz.]{{\expandafter\includegraphics\expandafter[\trimopt]{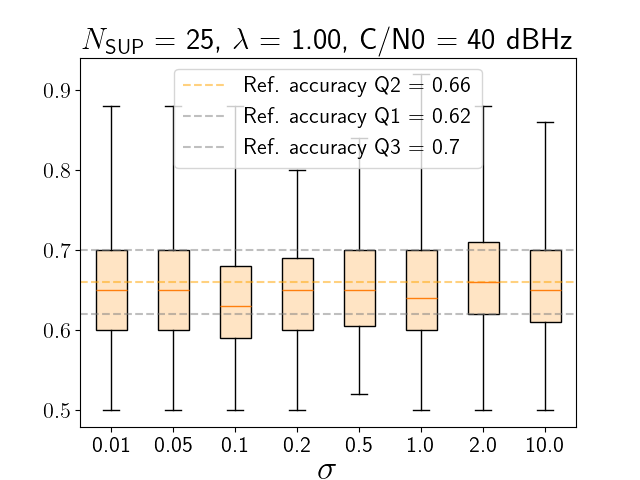} }}
\subfloat[\centering \CNZERO = 43 \dBHz.]{{\expandafter\includegraphics\expandafter[\trimopt]{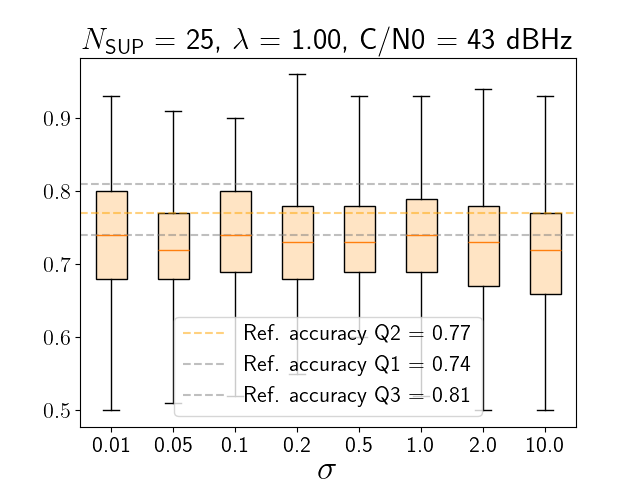} }}
\caption{Quartiles of the maximum accuracy for $N_\text{SUP} = 25$.}
\label{figure:25_cases}
\end{figure}

\end{document}